\def\year{2020}\relax

\documentclass[letterpaper]{article} 
\usepackage{aaai20}                  
\usepackage{times}                   
\usepackage{helvet}                  
\usepackage{courier}                 
\usepackage[hyphens]{url}            
\usepackage{graphicx}                
\usepackage{graphicx}                
\usepackage{color, colortbl}
\usepackage[utf8]{inputenc}
\usepackage{quoting}
\usepackage{rotating}
\usepackage{makecell}
\usepackage{multirow}

\title{{Domain-Level Explainability -- A Challenge for Creating Trust in Superhuman AI Strategies}}
\author{
  \Large \textbf{Jonas Andrulis\textsuperscript{\rm 1}, Ole Meyer\textsuperscript{\rm 2},
  Gr\'egory Schott\textsuperscript{\rm 1}, Samuel Weinbach\textsuperscript{\rm 1}, Volker
  Gruhn\textsuperscript{\rm 2}} \\
  \textsuperscript{\rm 1}Aleph Alpha GmbH, Heidelberg, Germany \\
  \{firstname.lastname\}@aleph-alpha.de \\
  \textsuperscript{\rm 2}University of Duisburg-Essen, Schützenbahn 70, 45127 Essen,
  Germany \\
  \{firstname.lastname\}@uni-due.de
}

\begin{document}

\maketitle


\begin{abstract}

For complex strategic video games, intelligent systems based on Deep Reinforcement
Learning (DRL) have demonstrated an impressive ability to learn solutions that can go
beyond human capabilities.
While this might create new opportunities for the development of assistance systems with
ground-breaking functionalities, applying this technology to real-world problems carries
significant risks and therefore requires trust and transparency.
Compared to other AI systems complex superhuman strategies are non-intuitive and difficult
to explain. A representative empirical performance evaluation in real-world scenarios is
often impossible.
Explainable AI (XAI) has improved the transparency of modern AI systems through a variety
of measures, however, research has not yet provided solutions enabling domain-level
insights for expert users of DRL systems in strategic situations.
In this position paper, we discuss the existence of superhuman DRL-based strategies, their
properties, the requirements and challenges for applying them to real-world environments,
and the need for explainability at the domain-level as a key ingredient to enable trust.

\end{abstract}


\section{Introduction}

Deep Reinforcement Learning (DRL) is an area of machine learning where the system learns
from interacting with the environment and actions are reinforced based on reward values.
The algorithms optimize the expected long-term reward and continuously improve their
actions and policies.
Reward signals are available in many scenarios and often come with lower costs than labels
required for supervised methods. This approach can even be used when no clear labeling is
possible \cite{feng2018reinforcement}.
Because DRL only specifies the problem to solve and not the solution, DRL systems have the
potential to achieve performance beyond that of domain experts. This superhuman potential
makes them interesting in complex and stragetic real-world challenges.

In strategic problems, an agent has to achieve a long-term objective through a complex set
of highly-significant actions.
Such problems are "dynamic, hostile, and smart" \cite{buro2003real} and share aspects of
complexity with video games, such as: decision under uncertainty, spatial and temporal
reasoning, and agent collaboration.
Strategy (video) games have been used to support real-world (military)
training \cite{Herz2002ComputerGA} and have also proven ideal for the development of DRL
AI.

In 2016, AlphaGo \cite{2016Natur.529..484S}, a Deep Reinforcement Learning (DRL)
algorithm, demonstrated a performance that surpassed that of the best human players of Go,
a strategy game considered beyond the reach of traditional AI due to its prohibitively
large branching factor.
One year later, AlphaZero \cite{2017Natur.550..354S} proved that an AI can learn a
superior Go strategy from the game's rules alone, without any expert input. Since then,
DeepMind's AlphaStar \cite{2019Natur.575..350V} and OpenAI Five \cite{openai2019dota} have
further advanced the capabilities of DRL systems.
This research did not only focus on existing games but introduced flexible game-like
environments for general strategic AI research \cite{tian2017elf}.

While the research on DRL-based strategies is steadily making progress for complex video
games, its transfer to practical real-world applications, where DRL may have a potential
for superhuman disruption, is lacking.
One reason may be found in low degrees of explainability which counteract human
understanding and acceptance. Actions formulated by an assistant system are only
implicitly learned, evaluated and encoded in a Deep Neural Network. Resulting strategies
cannot be represented with regular planning techniques and explicit explanations are not
easily available. We argue new forms of explainability need to be developed.

In this paper, we discuss the potential for superhuman strategies and possible challenges
based on our combined experience from industrial and practical uses.
From a combination of machine learning, trust, and innovation research we identify
domain-level transparency as one of the core difficulties facing AI applications in order
to leverage successful superhuman DRL solutions in strategic real-world scenarios and
specify approaches to provide explainability.


\newcommand\RotatedTexte[1]{\rotatebox{90}{\parbox{2cm}{\centering#1}}}
\newcommand\RotatedText[1]{\rotatebox[origin=c]{90}{#1}}

\setcellgapes{1pt}

\begin{table*}[!ht]
  \centering
  \caption{Explainability Challenges for Superhuman Strategies found by DRL Agents}
  \smallskip
  \makegapedcells

\begin{tabular}{|p{0.22cm}|p{0.6cm}|c|p{14.6cm}|}
  \hline
  \parbox[t]{0mm}{\multirow{21}{*}{\RotatedText{\bf Domain Level}}}

  & \parbox[t]{3mm}{\multirow{14}{*}{\RotatedTexte{\bf Strategic Complexity}}}

    & {\bf C1} & {\bf Spatial \& temporal reasoning:}
    Actions are not only conditioned on the currently “visible” state but also on past and
    future states. Relevance of observations depends on both state and actions.
    \\ \cline{3-4}

    & & {\bf C2} & {\bf Collaboration:}
    Interdependencies between actions of competitors and collaborators are essential in
    game-theory-like scenarios.
    \\ \cline{3-4}

    & & {\bf C3} & \makecell[l]{{\bf Decision-making under uncertainty:}
    Incomplete and uncertain information plays a major role in \\
    detemining an optimal strategy.}
    \\ \cline{3-4}

    & & {\bf C4} & {\bf Resource management:}
    Short term resources must be allocated towards a long-term strategic goal. Measurable
    advantages may appear long after a decisive action, with results appearing
    disadvantageous in the mean time.
    \\ \cline{3-4}
    
    & & {\bf C5} & {\bf Opponent modeling \& learning:}
    Learning from experience and adapting to scenarios and opponents.
    \\ \cline{3-4}
    
    & & {\bf C6} & {\bf Adversarial real-time planning:}
    Long-time planning may be required due to sparse reward signals.
    \\ \cline{3-4}
    
    & & {\bf C7} & {\bf Huge action- \& state-spaces:} 
    Some environments have a number of variables, observations, possible actions, or rules
    that is much larger than in classical strategy games.
    \\ \cline{3-4}
    
  \cline{2-4}
  & \parbox[t]{2mm}{\multirow{7}{*}{\RotatedTexte{\bf Strategic\\ Explainability}}}
  
    & {\bf E1} & {\bf Future projections:}
    Analysis of (potential) future states, events, and competitor behavior.
    \\ \cline{3-4}

    & & {\bf E2} & {\bf Hypotheticals scenarios:}
    Study of ``what-if'' scenarios on changes/hypothetical/potential future states.
    \\ \cline{3-4}
    
    & & {\bf E3} & \makecell[l]{{\bf Risk, transparency \& safety:}
    Risks due to “real” randomness or uncertain collaborator/competitor \\ behavior.}
    \\ \cline{3-4}

    & & {\bf E4} & {\bf Uncertainty:}
    Simulation results have to indicate how well the model is likely to capture a given
    situation (e.g. detection of out-of-distribution cases).
    \\ \cline{3-4}

 \hline
 \end{tabular}
 \label{table:requirements}
\end{table*}


\section{Potential barriers and challenges}

Strategies adopted by DRL agents are similar to disruptive innovation processes in
business models \cite{christensen2013innovator}, as they are able to provide superhuman
solutions that challenge established structures and theories.
Following the description of innovative business models \cite{chesbrough2010business}, we
derive preconditions for superhuman strategic disruptions based on DRL-AI, which are:
existence, potential and trust.

In following sections, we examine each of these three requirements based on existing
use-cases, applications, and research examples.
We discuss potential challenges for superhuman strategies of DRL agents and provide an
outlook on the critical questions to be addressed in order to enable the use of DRL
applications in industrial practice, especially by creating trust and enhancing
domain-level explainability.


\section{Existence of strategic real-world challenges}

Strategic environments with high complexity and high uncertainty have been traditionally
tackled with simplified (stochastic) models or scenario
planning \cite{schoemaker2004forecasting}.
Academic strategic models along with application of game theory to conflict
research \cite{ConflictStudies2017} and operations research \cite{MilitaryOR} began with
World War II and grew in popularity during Cold War conflicts.
Market competition has since gained a considerable
importance \cite{moorthy1993competitive} and these methods have proven helpful in price
strategy \cite{pricingstrategy} and logistics \cite{cachon2006game} problems.

The scope of problems that can be addressed this way is, however, limited:
in Table~\ref{table:requirements}, we compile strategic complexity challenges that can be
seen in research \cite{buro2003real} and applications (labeled C1 to C7 in the table).
When some of these criteria are met, simplification is required to approach the problem
with established strategic planning methods while DRL agents have been shown to master
this level of strategic complexity.

For example, OpenAI Five \cite{openai2019dota} is a remarkable case that excels in all
those criteria:
in DOTA 2, an incomplete information game (C3), teams of five compete against each other
(C2) with limited information about the actions of the competitors (C5).
Successful moves require coordination and planning (C1), forcing players to adapt and
build ressources in a long-term effort (C4).
In a match, each agent plans up to 80.000 turns with each time up to 170.000 possible
actions (C7) while there are no meaningful rewards until a match is either won or lost.
OpenAI has been able to build agents that compete successfully against the world's best
DOTA 2 players in real-time competition (C6).

Strategic problems with complexity beyond the limits of established strategy models and
scenario planning can be found in real-world areas and there are indications that DRL may
be uniquely suited to address these challenges.


\section{Potential for superhuman disruption}

In competitions against world's best human players, DRL has been able to achieve
unexpected and superior results -- which we classify as superhuman disruptions.
This has even been the case for strategy games that have received the attention of
millions of players for decades or even centuries.

In Go, one great example for this is turn 37 of game 2 between
AlphaGo \cite{2016Natur.529..484S} and Lee Sedol that resulted in the AI's
victory \cite{Turn37}.
\begin{quoting}[begintext={``},
endtext={''\footnote{https://deepmind.com/research/case-studies/alphago-the-story-so-far}}]
  During the games, AlphaGo played several inventive winning moves, several of which --
  including move 37 in game two -- were so surprising that they upended hundreds of years
  of wisdom. Players of all levels have extensively examined these moves ever since.
\end{quoting}

Similar results have been observed in DOTA 2 -- which is massively more complex than Go
and reaching the level of complexity of real-world scenarios -- with one of the world's
best players stating after losing against OpenAI Five:
\begin{quoting}[begintext={``},
endtext={''\footnote{William “BlitzDota” Lee on OpenAI Five playing DOTA 2.}}]
  It did things that we had never seen anybody else do and it has set a type of play style
  that we pretty much just copy now. When I see the bot make a play, it clicks in my head.
  I'm like, `why aren't we doing that?'
\end{quoting}

The existance of duperhuman disruptive results for any given problem can never be
guaranteed.
We argue, however, that experience with complex games has shown that we can expect
superhuman disruptions in real applications as well.
The reason for the lack of superhuman DRL in real-world scenarios cannot be attributed to
a lack of potential.


\section{Trust in superhuman AI-strategies}

Even if technical challenges can be solved, trust is essential in practical AI
applications \cite{2019trustinai}, as safety requirements and threats can lead to economic
costs, risks, and even regulatory issues. A human expert has to make the decision to
delegate to the AI some aspect of importance in achieving a goal without the possibility
to completely verify the AI's suggestion and all its potential
implications \cite{grodzinsky2011developing}.

Four key components have been shown to build trust in AI: transparency, reliability,
tangibility and task characteristrics (Glikson and Woolley 2020). Of those four,
transparency and reliability do not depend on the specific system. They can be the basis
for general machine learning requirements and correspond to the technically researched
areas of robustness and explainability. Both define a trusted zone shown on Figure 1.

This Figure illustrate that trust might be obtained for robust and not explainable models,
or explainable but not robust ones, while higher trust is achieved for models that are
both robust and explainable.

\paragraph{Robustness:}
Robustness is a concept developed in control theory \cite{sastry2011adaptive}, which is
intended for dealing with the effects of uncertainties.  This idea has been applied to
machine learning models by measuring the impact of fluctuating inputs or environments such
as uncertainties coming from modeling errors \cite{reinelt2002comparing}, poor
generalization due to overfitting, or intentional adversial attacks.  Robustness is an
ongoing challenge for DRL.  It has been very difficult to build generalizing DRL
agents \cite{cobbe2019quantifying} with recent success only for simple
environments \cite{badia2020agent57} and, for some DRL models, even naively executed
adversial attacks can have a significant impact on
performance \cite{pattanaik2017robust}. Seemingly unimportant changes in hyperparameters
or the random seed can also produce drastically different
results \cite{henderson2017deep}.  Given these issues with robustness, one way to satisfy
safety concerns is through the integration of safety
constraints \cite{junges2016safety,cheng2019end}.  Overall, research in this area is still
active and currently represents a major challenge in the development of DRL systems.

\paragraph{Explainability:}
Current AI explainability methods can be divided into three categories: model
explainability, outcome explainability and model inspection \cite{guidotti2018survey}.

\textit{Model explainability} approaches XAI by approximating the results of one model
with a second model that is by design easier to understand -- for example, by using decision
trees \cite{van_der_waa_contrastive_2018,johansson_evolving_2009,craven_extracting_1996}.
However, even human-readable rules quickly become complex and incomprehensible, especially
when they occur in large numbers \cite{lage2019evaluation}.

\textit{Outcome explainability} approaches XAI by illustrating the effects of different
inputs on the outputs of the model while mainly ignoring what is going on within the model
itself.  The most common technique masks the actual input spotting the input-subset
primarily responsible for the model result.  The layer-wise relevance propagation
approach \cite{bach_pixel-wise_2015} uses backpropagation.  Creating attention maps over
the input is another possibility \cite{xu_show_2015,fong_interpretable_2017}.  There are
also experiments of model agnostic methods, such as Local Interpretable Model-Agnostic
Explanations (LIME) \cite{ribeiro_why_2016}.  This approach is particularly well suited
for images, because in this case humans are able to process the large amount of
information (pixel by pixel) quickly.  In other areas, however, this becomes difficult
because the ability to capture the information is not sufficiently available.

\textit{Model inspection} directly analyzes on a technical level how model results are
generated. Examples are sensitivity analysis \cite{saltelli_sensitivity_2002}, the
representation of dependencies between features and outputs \cite{friedman_greedy_2001}
and Activation Maximization (AM) \cite{yosinski_understanding_2015}.

According to \cite{puiutta2020explainable} which reviews XAI methods designed for DRL
systems, those can be categorized into two major groups: scope of explaination and time of
information extraction.

Scope of explanation can be either global, when the structure of the model is transparent
to the user, or local, when user can access the explanations for a specific decision of
the model. It has however been noted that `human users tend to favor explanations about
policy rather than about single actions' \cite{van_der_waa_contrastive_2018}.

Depending on the time when the explanation is produced it can be either intrinstic or
post-hoc.
Simple intrinstic interpretability, produced by self-explanatory approximators for the
policy function at the time of
training \cite{hein2018interpretable,verma2018programmatically}, suffers from low
prediction performance.
A more common approach \cite{liu2018toward,madumal2019explainable} is with post-hoc
explanation of models. Those are constructed after training by creating a simpler
model \cite{Du_2019} that provides acurate explanations which can however be difficult to
interpret.
It seems unlikely that superhuman strategies would be meaningfully preserved in simplified
explanation models.

In their review, the authors of \cite{puiutta2020explainable} find that XAI for
reinforcement learning needs to exhibit context awareness by adapting to environment and
user. One of the established and easily understandable ways to do this is to offer
contrastive explanations comparing different strategy options. This kind of XAI output -
which is especially useful for domain experts without any further AI knowledge - can be
found in three of thirteen papers they reviewed:
\cite{madumal2019explainable,sequeira2019interestingness,van_der_waa_contrastive_2018}.

As key finding, they conclude that the ability to not only extract or generate
explanations for the decisions of the model, but also to present this information in a way
that is understandable by human (non-expert) users, makes it possible to predict the
behaviour of a model.
This definition of XAI implicitly assumes that the expert perfectly understands the
strategic problem and can easily judge the right action.
In a domain of high complexity and uncertainty, where intuitive judgment should not be
trusted \cite{hogarth2001educating}, this cannot be easily expected and strategic
explanations become a key challenge -- making AlphaGo's turn 37 in game 2 predictable
through XAI is a far greater challenge than the examples the authors had in mind.

\paragraph{Summary:}
Trust is an essential factor for the deployment and leveraging of DRL systems in
real-world scenarios. Both current robustness and explainability methods are not suited
for the requirements of complex strategic environments and constitute areas where further
research is required.


\section{Domain-Level XAI}

Considering the limits for XAI in strategy contexts, discussed in the last section, one
may ask: What are the key ingredients for a 'strategic' XAI that will help human experts
learn from superhuman AIs? This question cannot be answered on a technical level alone but
must also address the strategic complexity (C1-C7) in a way that a domain expert with no
machine learning mastery can use this information.

To achieve this, established tools for scenario planning \cite{schoemaker2004forecasting}
can be adopted and translated into requirements for 'strategic explainability' in a DRL
context (E1-E4 in Table~\ref{table:requirements}).  While scenario planning is limited and
mostly qualitative, its approach to uncertainty \cite{courtney1997strategy} and the
criteria for scenario selection provide solid foundations for strategic explainability of
DRL-AI results.

The main questions and drivers for scenarios are predictions of the future (E1) and the
impact of changes in hypothetical scenarios (E2).  While classical scenario planning has
no way of quantify probability distributions, DRL adds this quantitative dimension with
the potential to add a measure for risk (E3) and model uncertainty (E4).

Transparency of superhuman AI-strategies is essential and future research must further
focus on domain-level explainability (E1-E4) in strategically complex environments
(C1-C7).
Established approaches to strategy modeling, such as scenario planning, may be a great
resource for building strategic AI-assistant systems and gaining the trust of experts in
the potential for superhuman disruptions.


\begin{figure}
  \centering
  \includegraphics[width=.475\textwidth]{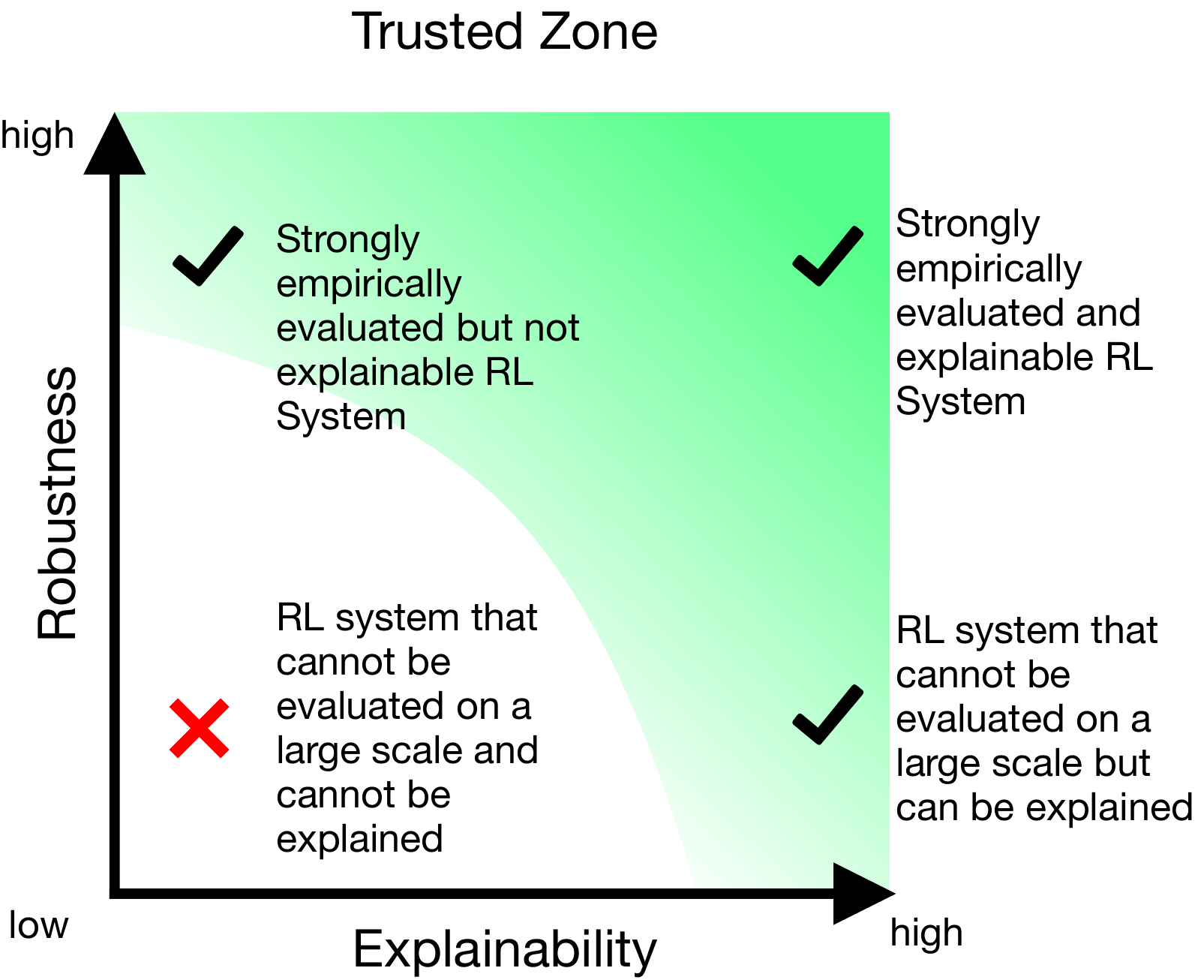}
  \caption{Trusted Zone for Strategic Real-World AI Systems}
  \label{fig:trusted_rl}
\end{figure}


\section{Conclusion \& Future Work}

There are real-world strategic use-cases that offer great potential for superhuman
innovation through the use of DRL-AIs.
Because the robustness of complex real-world strategies often cannot be empirically
validated, trust in these systems must be build through transparency and explainability.
Current XAI methods cannot offer the domain-level strategy explanations that are necessary
for an expert to understand counter-intuitive superhuman
strategies \cite{hogarth2001educating}.
In order to build trust-enabling transparency into strategy AIs, the current concepts of
explainability need to be enhanced.
The implicitly learned strategic complexity requires an explainability that can address
concepts beyond the relationship of individual input and output combinations for users
without technical machine learning knowledge.
Those need to be implemented as readily available tools that can be applied to AI agents.
Finally, future studies will have to show that domain-level strategic explainability is
possible so that human experts can trust and benefit from superhuman strategies issued by
a DRL-AI in real-world applications.


\bibliographystyle{aaai}
\fontsize{9.0pt}{10.0pt} \selectfont
\bibliography{bibliography.bib}

\end{document}